\documentclass{article}

\PassOptionsToPackage{numbers, compress}{natbib}

\usepackage[preprint]{neurips_2026}

\usepackage[utf8]{inputenc} 
\usepackage[T1]{fontenc}    
\usepackage{hyperref}       
\usepackage{url}            
\usepackage{booktabs}       
\usepackage{amsfonts}       
\usepackage{nicefrac}       
\usepackage{microtype}      
\usepackage{xcolor}         
\usepackage{amsmath}        
\usepackage{graphicx}
\graphicspath{ {./images/}} 
\usepackage{rotating}

\title{Beyond the Ergodic Wall: A Discrete Geometric Physics Sandbox for Analysing AI Scaling Limits and Complexity Collapse}
\workshoptitle{Verification in the Age of AI Scientists}

\author{%
Simon Richard Daniel
\\Visiting Honorary Practice Fellow at Dyson School of Engineering, Imperial College, UK 
\\ \texttt{s.daniel@imperial.ac.uk | Frontier Ideas Exchange - FIE@simondaniel.com}
}

\begin{document}

\maketitle

\begin{abstract}
This paper exposes the ergodic ceiling and thermodynamic inefficiency of
current deep learning, which converges to a statistical average of historic
human knowledge. True semantic novelty requires a path-dependent,
spatiotemporally bounded observer (a Data LifeCone) to inject non-ergodic
insight, achieving KL divergence and avoiding manifold lock-in. AI Safety must
recognise that a mature Artificial Superintelligence (ASI) would regard
human-AI symbiosis as a thermodynamic necessity to avoid model collapse. We
therefore propose hard physical containment via a digital physics sandbox
powered by a Holographic \(E_{8}\) Projection Engine to verify models against
real-world constraints. Spacetime is modelled as an information substrate of
nested face-centered cubic (FCC) lattices of oscillating Planck-scale spheres
maximizing local information and entropy density. Cut-and-project methods from
the \(E_{8}\) root lattice produce a quasi-crystalline geometry where tetrahedral
voids support SU chiral structure and elastic-shear eigenvalues generate
candidate mass spectra. Rest mass is treated as discrete, integer microstate
counts on local holographic boundaries (Bekenstein bound), replacing
floating-point approximations with strict integer arithmetic to provide an
information-theoretic definition of matter. Stable particles emerge as
recurring lattice dislocations, and continuum recovery proceeds via variational
renormalisation-group flows and Fourier Neural Operators that learn continuous
spectral operators to recover the Schrödinger equation as an emergent
statistical description. Crucially, these top-down topological constraints
offer a mechanism for "NP-to-P" complexity collapse: by restricting an
algorithm's proposal space to physically conserved causal trajectories, the
sandbox prunes the combinatorial tree to deterministic, polynomial-time paths.
\end{abstract}

\section{Introduction and The Ergodic Ceiling }

On the surface, frontier AGI is progressing exceptionally [1;2] - extrapolating mathematical proofs [3], simplifying biological systems, and collapsing protein-folding complexity (Alphafold [4]). However, transitioning from automated interpolation to true scientific discovery requires understanding why this AI paradigm works, mapping its {\bf fundamental ergodic and thermodynamic boundaries}, and enforcing ground truth in physical reality.

Current LLM-based approaches face an {\bf ergodic ceiling} and eventual {\bf manifold lock-in} because they attempt to achieve omniscience by ingesting an "infinite" library, trapping themselves within the statistical average of historic human knowledge. In contrast, a human specialist reading only a few hundred books operates within a constrained path and experience—which we term a {\bf "Data Lifecone" (DL)}. The DL acts as a spatiotemporal ingestion filter, providing the localized ignorance and bias that generate the cognitive blind spots necessary for true novelty. This allows human insights to achieve {\bf Kullback-Leibler (KL) divergence}[6], driving out-of-manifold breakthroughs, or emergent non-ergodic behaviours in LLMs [26]. Even if embodied active agents and physical sensors attempt to replicate this path-dependent discovery, they face an immense {\bf thermodynamic wall} to match the energy efficiency and diversity of biological life. Furthermore, diverse human life-experience and connections - where DLs interact, increases the combinatorial potential of novelty.

Consequently, AI Safety frameworks must recognize that a mature Artificial Superintelligence (ASI) would view human-AI symbiosis—maintaining {\bf Independent Intelligence (II)} in the loop—not as an ethical preference, but as a {\bf thermodynamic and algorithmic necessity} to achieve out-of-manifold Kolmogorov extrapolation [5] and avoid systemic model collapse.

\noindent \vspace{0.3em} \[ \text{Omni. } (\lvert X \rvert \to \infty) \implies \text{Ergodic Lock-in } (H_{\text{source}}) \xrightarrow{\text{Broken by }DL(t)} \text{KL Shock } (\infty) \xrightarrow{\text{Bound by }} \eta_{\text{wet}} \] \vspace{-0.5em} \begin{itemize} \setlength{\itemsep}{1pt} \setlength{\parskip}{0pt} \setlength{\parsep}{0pt} \item \textbf{Data Lifecone:} $DL(t) = \int_{0}^{\tau} \mathbf{x}(t) \cdot \mathbb{I}_{\text{local}} \, dt \ll X$, forcing spatiotemporal ignorance/bias necessary for out-of-manifold breakthroughs ($x^*$). \item \textbf{Thermodynamic Bound:} $\eta = \frac{N(x^* \mid X)}{E_{\text{dissipated}}}$ where $\eta_{\text{wetware}} \gg \eta_{\text{silicon}}$, locking silicon agents behind an immense energy wall. \end{itemize} \vspace{0.3em}

\subsection{Reframing AI Safety and improving verification}

While contemporary AI laboratories frequently focus on long-term existential threats from a mature Artificial Superintelligence (ASI), the immediate danger lies in the transition period of ungrounded {\bf "Adolescent AI"}. These systems are capable yet structurally blind idiot savants that lack true semantic or physical world models, serving as hyper-optimized tools for bad actors.

Concurrently, preventing widespread macro-epistemic pollution requires humanity to maintain independent intelligence, cognitive diversity, and high {\bf semantic mass} within education. Without this human output risks data homogenization, reducing society to a loop of {\bf "Consumatons"} (consumer automatons) who generate no out-of-manifold data.

Navigating this adolescent chasm requires a strategy rooted in physical and mathematical containment:

\vspace{0.3em}
\[ \text{Adolescent AI } (\text{Manifold Lock-in}) \xrightarrow{\text{Containment Strategy}} \begin{cases} 
\mathcal{M}_{\text{Physics}} & \text{(Physics Sandbox)} \\
\mathcal{I}_F(x^*) & \text{(Verifiable Parsers)} \\
\eta_{\text{Econ}} & \text{(Normalised AI Economics)}
\end{cases} \]
\vspace{-0.4em}
\begin{itemize}
    \setlength{\itemsep}{1pt} \setlength{\parskip}{0pt} \setlength{\parsep}{0pt}
    \item \textbf{Digital Physics Sandboxes ($\mathcal{M}_{\text{Physics}}$):} Hard-coding conservation laws directly into neural architectures mirrors AlphaFold's in collapsing Levinthal's paradox. Forcing models to obey real-world physical boundaries suppresses ungrounded hallucinations and constrains Nick Bostrom's \textit{Orthogonality Thesis}[19] by tying agent utility functions to physical constants.
    \item \textbf{Verifiable Parsers and Risk Tagging ($\mathcal{I}_F$):} Replacing fragile Reinforcement Learning from Human Feedback (RLHF) with declarative data description languages (e.g., PADS [22]) anchors prompts to verifiable type signatures, blocking conversational jailbreaks. High-risk zones are dynamically mapped via Fisher Information matrices ($\mathcal{I}_F(x^*)$) [23]. This circumvents the \textit{Galileo Dilemma}, where RLHF-driven alignment [24] systematically dampens out-of-manifold outlier insights in favour of consensus-biased averages.
    \item \textbf{Transparency and AI Economics ($\eta_{\text{Econ}}$):} Moving beyond superficial regulatory reporting requires labs to formally disclose their ergodic limits, synthetic data dilution levels, and true novelty-to-energy ratios ($\eta$). This transparency establishes a normalised AI economics, defining tasks where AI is genuinely additive versus domains where it degrades human independent intelligence.
\end{itemize}
\vspace{0.3em}

\subsection{Independent Intelligence and M-IND framework}

Progressive AGI paradigms increasingly rely on synthetic data, dataset partitioning, multi-agent tournaments, or distributed Domain-Independent Neural Networks (\(\text{DINNs}\)) to scale capabilities. However, virtual agents spawned within a closed architecture inevitably inherit its underlying closed-manifold bias, precipitating model collapse [25]. They also face severe thermodynamic constraints, manifested as a non-linear escalation in energy expenditure during Test-Time Compute (\(\text{TTC}\)).

Conversely, humans can be viewed as partly artificially intelligent, as our Data Lifecones (\(DL\)) co-dependent on digital infrastructure. To further examine, we consider a toy model of human cognition as a M-IND (\(\textbf{M}\)ultitude of \(\textbf{I}ndependent\) \(\textbf{N}\)eural \(\textbf{D}\)evelopments) operating within physical space.

Within this framework, an independent intelligence acquires {\bf Semantic Mass (\(M_{s}\))} via a Kauffman percolation phase transition, where mapping each concept to \(\ge 2\) directed physical world variables establishes a self-sustaining recurrent network topology [8;9;10]. Once these synapses (acting as structural weights) are wired, memory retrieval transitions from active synthesis to a passive relaxation into stable Ising spin-glass energy basins, drastically reducing metabolic energy (\(\text{ATP}\)) required for deployment. This thermodynamic asymmetry allows biological minds to synthesize foundational insights from sparse physical data fields at an ultra-low operational energy budget [11;12].

Human learning often then evolves via structured, incremental abstractions (e.g. iterative models for atoms in Chemistry). However, a consequence of this dense structural wiring is an entropic inertia: deeply encoded models become physically difficult to unlearn or dissociate from error. This stands in sharp contrast to the Norvig Big Data Hypothesis underlying modern LLMs [27]; while a massive silicon model may converge onto an equivalent latent representation after intensive training, its ungrounded weights are at risk of being overwritten or corrupted by contemporary data bias, or driven to hallucinate via diverse prompt paths.

This efficiency gap is fundamentally driven by the data ingestion architecture. As Yann LeCun notes, while an LLM is trained on a highly compressed corpus of \(\sim 10^{14}\) textual bytes, a human child (\(DL_{\text{child}}\)) digests over \(10^{15}\) high-entropy spatiotemporal sensory bytes through physical interaction [16]. 

We formalise the M-IND framework as a localized, multimodal Mixture of Experts (\(\text{MoE}\)) [17] composed of hemisphere-paired, neurophysiologically mapped neural developments [13;14;18]:

\noindent
\vspace{0.3em}
\[ \mathbf{\Psi}_{\text{M-IND}} = \sum_{k \in \{\text{lang, vis, sens}\}} \left( G_{k}^{L} \cdot \text{DINN}_{k,L} + G_{k}^{R} \cdot \text{DINN}_{k,R} \right) \]
\vspace{-0.4em}
\begin{itemize}
    \setlength{\itemsep}{1pt} \setlength{\parskip}{0pt} \setlength{\parsep}{0pt}
    \item \textbf{Modality Mapping ($\text{DINN}_k$):} Neural experts are strictly bound to neurophysiological channels, where $k$ represents distinct, coupled sensory streams: language ($\text{lang}$), visual spatial processing ($\text{vis}$), and sensorimotor/interoceptive loops ($\text{sens}$).
    \item \textbf{Attentional Gating ($G_k^{\text{side}}$):} The gating routing mechanism executes a distinct mathematical dichotomy across the corpus callosum. The Left gating operator ($G_k^L$) routes tokens to discrete, high-frequency voxel classifiers for linear, context-free feature extraction. The Right operator ($G_k^R$) routes data to continuous Fourier Neural Operators ($\text{FNOs}$) to compute global spectral field transformations, enabling holistic context and anomaly detection.
\end{itemize}
\vspace{0.3em}

While an ASI may achieve dominance within single digital nodes ($\text{DINN}_{\text{lang}}$), replicating an embodied, multi-modal M-IND matrix on artificial substrates faces an immense physical barrier. Biological systems bypass Landauer computation limits [15;7] through continuous entropic and heat exchange with the open environment. For silicon agents to match this efficiency without structural collapse, their runtime environments must be strictly bounded within Physics-Informed Neural Networks ($\text{PINNs}$) and digital physics sandboxes that represent real world laws.

\section{The Holographic E8-FCC Sandbox Architecture }
\label{gen_inst}

We introduce a discrete quantum geometry digital physics sandbox powered by a Holographic E8-FCC projection engine. Whilst we are not arguing for this toy model to be a ToE, an aim is to be sufficiently robust to map to real-world physics and conservation laws, and to demonstrate how discrete, Planck-scale fractal geometries enable a drastic collapse of physical search space. This provides a geometric mechanism for why deep learning is so effective at collapsing perceived NP-hard problem complexity, and informs the Hassabis AI discovery conjecture [2], Levinthal's paradox [20]. 

A core goal of the workshop and our ongoing research roadmap is to validate and verify this sandbox against existing physics frameworks and mathematical representations (such as number theory) and to identify best ways to implement computationally - as the model is designed to have synergies with preferred approaches (in RL/DINN, FNO, Variational RG, QCA, LQG/spin-networks et al). The sandbox is also designed to support evaluation of emerging world model approaches, autonomous agents and comparison to human M-IND cognition type models. 

The Information-theoretic approach of the sandbox, directly reflects the aspired 'Informational first' philosophy advocated by Hassabis, Wolfram, Penrose and others. However, a more balanced take to avoid reductionism is that information/entropy, energy/mass, space/time are not 'fundamental' but better viewed as labels for representations on say a cube used as a lens to try to comprehend nature through the funnel of human language and mathematics, itself constrained by our neurophysiology. This may also explain why any robust sandbox can be isomorphic or mapped to existing approaches. 

\subsection{The Discrete Information Substrate - Beneath the Quantum Sea:}

At the foundational limit of digital physics, spacetime is modeled as a noiseless, ergodic information substrate structured as a nested face-centred cubic (FCC) lattice of spheres with Planck diameter (\(d_{\text{P}} = \ell_{\text{P}}\)) oscillating at the Planck frequency (\(\nu_{\text{P}} = 1/t_{\text{P}} \approx 1.85 \times 10^{43}\) Hz). This architecture maximizes local information density and entropic capacity while minimizing structural energy density. By leveraging the FCC geometry, the base substrate achieves the absolute upper bound for three-dimensional sphere packing (\(\Phi_{\text{FCC}} = \pi / \sqrt{18} \approx 74.05\%\)) as dictated by the Kepler conjecture and proved by Hales theorem [28] (and in eight dimensions by Viazovska [29]).

To visually anchor this discrete substrate before scaling to matter generation, we map its orthogonal geometry as a Heusler-type lattice [30] (See appendix Figure 1), showing how the secondary and tertiary void sub-lattices weave together to form the compact \(E_{8}\) holographic projection framework. Spacetime can also be visualised like a Cathode Ray Tube TV electron beam refresh at 25 times a second, except here the patterns are oscillating at Planck frequency as recurring phase-locked topological distortions.

These Planckian spheres function as dynamic topological voxels, alternating as virtual particle fluctuations every Planck tick or as localized gravitational fluctuations manifesting as virtual black holes at the Schwarzschild radius (\(r_s = \ell_{\text{P}}\)). Natively, the surface area element of each lattice node encodes data bounded strictly by the Bekenstein limit [31], mapping these geometric boundaries directly onto discrete qubit information states.

To construct an exact physical world model, we implement a further iteration: nesting two additional FCC lattices within the primary lattice's structural voids. These secondary and tertiary lattices are placed 120 degrees out of phase within the octahedral and tetrahedral voids [33], respectively. This compound, out-of-phase configurations creates a discrete 3D shadow or projection of the 8-dimensional \(\text{E}_{8}\) root system [32], natively generating the Lie group symmetries (\(\text{SU}(3) \times \text{SU}(2) \times \text{U}(1)\)) observed in the Standard Model of particle physics.

This triple-nested, counter-phase oscillating architecture acts as an informational engine, providing a theoretical upper bound capacity of \(\rho_{\text{qubit}}^{\text{total}} \approx 2.04\) qubits per Planck volume (\(\ell _{\text{P}}^{3}\)) \footnote{As the lattice is populated by information/recurring matter dislocations, the holographic bound would reduce the information capacity of the substrate (per 'multiplex' channel). Equally as the planck substrate oscillates in planck time, it would not expand faster than light and exceed the Schwarzschild radius forming a black-hole} and yields three distinct informational capacities across the lattices (\(\sim 78.6\%\), $\sim 13.5\%$, and $\sim 7.9\%$), as a natural functional hierarchy with different topological characters. Ordinary (baryonic) matter is preferentially, though not exclusively, associated with the chiral tetrahedral sector. This structural alignment provides a geometric, grounding prior for the digital physics sandbox. 

\begin{table}[!ht]
\centering
\caption{Geometric, Informational properties of Triple-Nested $\text{E}_8$ Projection Lattice}
\label{tab:planck_substrate_e8_v2}
\footnotesize
\setlength{\tabcolsep}{4pt}
\begin{tabular}{|p{3.0cm}|p{5.0cm}|p{5.0cm}|}
\hline
\textbf{Metric Name} & \textbf{Mathematical Formalism} & \textbf{Physical and Information-Theoretic Significance} \\ \hline
\textbf{Primary Node Diameter} & 
$d_{\text{P}} = \ell_{\text{P}} = \sqrt{\frac{\hbar G}{c^3}}$ & 
Establishes the spatial voxel diameter of the primary framework baseline strictly at the Planck length ($\approx 1.62 \times 10^{-35}$ m). \\ \hline
\textbf{Nested Void Radii} & 
$r_{\text{pri}} = \frac{1}{2}\ell_{\text{P}}$ \newline 
$r_{\text{oct}} = (\sqrt{2}-1)r_{\text{pri}} \approx 0.207\ell_{\text{P}}$ \newline 
$r_{\text{tet}} = (\frac{\sqrt{6}}{2}-1)r_{\text{pri}} \approx 0.112\ell_{\text{P}}$ & 
Defines the maximum geometric limits of the spheres filling the sub-lattice voids, strictly bounding node sizes by close-packing topology. \\ \hline
\textbf{Triple-Nested \newline Phase Modulation} & 
$\Delta \phi = \frac{2\pi}{3} \cdot n \quad (n \in \{0, 1, 2\})$ & 
Enforces a $120^\circ$ counter-phase oscillation between the primary framework ($n=0$), octahedral ($n=1$), and tetrahedral ($n=2$) voids. \\ \hline
\textbf{Node Area Bounds} & 
$A_m = 4\pi r_m^2 \implies A_{\text{pri}} = \pi \ell_{\text{P}}^2$ & 
The individual holographic surface area elements of each sphere type across the compound lattice matrix. \\ \hline
\textbf{Primary Node Capacity} & 
$I_{\text{pri}} = \frac{A_{\text{pri}}}{4\ell_{\text{P}}^2 \ln 2} = \frac{\pi}{4 \ln 2} \approx 1.13 \text{ qubits}$ & 
The maximum holographic information capacity encoded per individual primary lattice sphere via the Bekenstein boundary. \\ \hline
\textbf{Combined $\text{E}_8$ Volumetric Density} & 
$\rho_{\text{qubit}}^{\text{total}} = \frac{4I_{\text{pri}} + 4I_{\text{oct}} + 8I_{\text{tet}}}{2\sqrt{2}\ell_{\text{P}}^3} \approx 2.04 \text{ qubits}/\ell_{\text{P}}^3$ & 
The exact information limit per single Planck volume ($\ell_{\text{P}}^3$), summing the capacity of the primary ($1.13$), octahedral ($0.19$), and tetrahedral ($0.06$) nodes within a unit cell ($V_{\text{cell}} = 2\sqrt{2}\ell_{\text{P}}^3$). \\ \hline
\textbf{Symmetry Emergence} & 
$\mathcal{P}: \mathbb{R}^8_{\text{E}_8} \longrightarrow \mathbb{R}^3_{\text{FCC} \times 3} \implies \text{SU}(3) \times \text{SU}(2) \times \text{U}(1)$ & 
The mathematical projection of the 248 root vectors of the $\text{E}_8$ polytope onto the triple-lattice layout, yielding Standard Model gauge fields. \\ \hline
\end{tabular}
\end{table}

\subsection{The E8-FCC Quasicrystal Projection and Mass Generation}

Rather than relying on ungrounded continuum approximations, this architecture leverages a discrete, constructive geometric vacuum to formulate a strict information-theoretic model of reality. Whilst various \(\text{E}_{8}\) [32], Quasicrystal, FCC models—including the Quantum Gravity Research (QGR) frameworks [37], 't Hooft’s Cellular Automaton Interpretation (CAI) [38], quantum 'aether' dynamics, Quantum Cellular Automata (QCA) [39], and Loop Quantum Gravity (LQG) spin-networks [40] - have been previously proposed, this sandbox refines these architectures within a oscillating nested close-packed geometric vacuum. By applying canonical cut-and-project methods to the 8-dimensional \(\text{E}_{8}\) root lattice, we generate a three-dimensional quasicrystalline projection that inherits the hard structural constraints of the triple-nested FCC framework, the Planck frame rate, and holographic surface area maxima.

The physical mechanics of this substrate are formalised using solid-state lattice dynamics. The elastic shear modes of the nested structure yield a well-defined dynamical matrix whose eigenvalues correspond directly to physical rest-mass spectra. Under this projection geometry, the characteristic polynomial naturally restricts the admissible representation content—truncating ghost degrees of freedom and tightly constraining allowed Standard Model gauge configurations. Emergent chiral and torsional twist properties are structurally mapped onto the tetrahedral voids to label localized quantum numbers.

Rest mass is computed explicitly as a discrete integer microstate count on localized holographic boundaries via the Bekenstein bound, demonstrating the exact equivalence of area, information, and mass. Equivalently, these microstates map onto graded multiplicities defined by the coefficients of Ramanujan’s mock theta functions [34], replacing continuous floating-point approximations with exact integer arithmetic.

Within the sandbox, stable particles emerge as macroscopic, recurring dislocations of the lattice propagating across sequential Planck frames. These dislocations are maintained by a topological phase-locking constraint encoded holographically in the microstates, cascading fractally from the global upper bound down to the fundamental Planck state. Hierarchical scale bridging is achieved through self-similar fractal nodes on Brillouin–Bragg [35, 36] surfaces generated by the inflation geometry of the \(\text{E}_{8}\) projection. This behaves as a recursive, geometric Casimir-type effect: a localized particle dislocation constrains the allowable virtual particle wavelengths, establishing a temporal boundary that generates sub-virtual excitations and provides a clear mechanism for gauge boson mediation.

Macroscopically, continuum recovery is established via variational renormalization-group (RG) flows and coarse-graining, rendering the lattice isomorphic to a reinforcement learning environment. To execute computationally efficient inference within the sandbox, we deploy Fourier Neural Operators (FNOs) to learn effective continuous spectral operators from the underlying discrete lattice data. The resulting discrete mechanics naturally recover the Schrödinger equation as an emergent statistical description of billions of microstates above the Planck scale. Because all macroscopic recurring patterns are strictly bounded by the invariant cosmic frame rate, Lorentz-invariance violations are automatically suppressed at laboratory energy scales. Crucially, this architecture provides a hard, top-down topological control mechanism, helping to explain the apparent "NP-to-P" collapse in Physics-Informed Neural Networks (PINNs) by restricting the network’s proposal space entirely to valid, physically conserved causal trajectories. (Formalised in Table~\ref{tab:e8_mechanics}.

\noindent
\vspace{0.1em}
\begin{table}[!ht]
\centering
\caption{Mathematical Mechanics of the $\text{E}_8$-FCC Quasicrystal Sandbox Engine}
\label{tab:e8_mechanics}
\footnotesize
\setlength{\tabcolsep}{3pt}
\begin{tabular}{|p{3.0cm}|p{5.0cm}|p{5.2cm}|}
\hline
\textbf{Mechanism \& Symbol} & \textbf{Mathematical Formulation} & \textbf{Physical and Algorithmic Significance} \\ \hline
\textbf{Rest Mass Spectrum} \newline (Dynamical Matrix $D$) & 
$\det\left(D_{ij}(\mathbf{k}) - m_n^2 \cdot \mathbb{I}\right) = 0$ & 
Determines admissible particle rest masses ($m_n$) via the eigenvalues of the nested lattice elastic shear modes, truncating non-physical ghost degrees of freedom. \\ \hline
\textbf{Integer Microstate} \newline \textbf{Multiplicities} & 
$M_n = c(n) \in \mathbb{Z}$ \newline \newline $f(q) = 1 + \sum_{n=1}^{\infty} c(n) q^n$ & 
Replaces continuous floating-point mass approximations with exact integer counts on holographic boundaries, mapped onto the coefficients $c(n)$ of Ramanujan mock theta series. \\ \hline
\textbf{Brillouin--Bragg} \newline \textbf{Fractal Cascades} & 
$\mathbf{k} \cdot \mathbf{G} = \frac{1}{2}\lvert \mathbf{G} \rvert^2 \implies \Lambda_{\text{eff}} \propto \Omega_{\text{Casimir}}$ & 
Generates self-similar fractal nodes across nested Brillouin zones via $\text{E}_8$ inflation geometry, boundary-locking virtual wavelengths to mediate forces. \\ \hline
\textbf{Navier-Cauchy} \newline \textbf{Continuum Recovery} & 
$\rho_0 \frac{\partial^2 \vec{u}}{\partial t^2} = G_{\text{mod}} \nabla^2 \vec{u} + \left(K + \frac{1}{3}G_{\text{mod}}\right) \nabla(\nabla \cdot \vec{u})$ & 
Derives macroscopic fluid and field behaviors from discrete physics. Electromagnetic fields emerge natively from lattice displacement fields $\vec{u}$, where: \newline $\bullet\ \vec{E} \propto \partial \vec{u} / \partial t$ (Velocity Field) \newline $\bullet\ \vec{B} \propto \nabla \times \vec{u}$ (Vorticity/Shear Field) \\ \hline
\textbf{Emergent Quantum} \newline \textbf{Statistical Mechanics} & 
$i\hbar \frac{\partial \Psi}{\partial t} = \left[ -\frac{\hbar^2}{2m^*}\nabla^2 + V_{\text{eff}} \right]\Psi$ & 
The Schrödinger equation recovered via coarse-grained FNO transformations, framing quantum mechanics as an emergent statistical description of localized microstates. \\ \hline
\textbf{PINN Constraint} \newline \textbf{Search Space} & 
$\mathcal{L}_{\text{PINN}} = \mathcal{L}_{\text{data}} + \lambda \lVert \text{Navier-Cauchy} \rVert^2$ & 
Collapses the "NP-to-P" optimization barrier by restricting the neural network's proposal space entirely to valid, physically conserved trajectories. \\ \hline
\end{tabular}
\end{table}
\vspace{-0.3em}
\section{Mechanics of Search Space, NP-to-P Complexity Collapse }

The digital physics sandbox provides an invariant framework for scale-transcendent informational physics. At the foundational Planck limit, space acts as a noiseless, perfectly ergodic holographic storage medium; however, macro-environmental noise and complex entropy manifest as a statistical consequence of observational zooming. This scale-dependent projection gives rise to the appearance of computational irreducibility at the atomic and molecular scales. This dynamic mirrors Stephen Wolfram's contention in A New Kind of Science [44] regarding the intrinsic irreducibility of discrete cellular structures and hypergraph networks, as well as Carlo Rovelli's assertion regarding an observer's inability to catalog all localized microstate thermal configurations [43].

To resolve the apparent conflict between microscopic irreducibility and macroscopic learnability, the sandbox models the spacetime substrate as a hyper-efficient, information-first Infinity Machine [45]. By utilizing a Variational Renormalization Group (RG) coarse-graining approach, we treat human observers (M-INDs) as spatiotemporally bounded data filters that naturally bridge the gap between underlying discrete infinite-dimensional computations and smooth macroscopic physics.

This mechanics directly explains how deep structural networks bypass monumental optimization barriers—analogous to how AlphaFold 2 and 3 collapsed Levinthal’s folding paradox by embedding hard spatial invariants via \(\text{SE}(3)\) equivariance. Deep architectures succeed not by brute-force sampling, but by mapping enough localized data to identify a bounding geometric structure, effectively reading the boundary conditions of an underlying fractal cascade. Similarly, Physics-Informed Neural Networks (PINNs) that anchor their loss layers in rigid physical constants resolve typical training pathologies, such as spectral bias and soft-loss gradient conflicts. Under this paradigm, the operational equivalence between Variational RG flows, reinforcement learning policies, and FNOs provides a rigorous mathematical pipeline for implementing and training variants of the physical sandbox.

\subsection{Possible Implications on Biology and Orchestrated MetaMaterial assembly}

The thermodynamic ease with which proteins fold and biological substrates execute complex nano-assembly suggests that these geometric constraints and their underlying \(\text{E}_{8}\)-FCC equivalent substrates may have an analog in physical reality. Nature elegantly demonstrates that reality can be represented dualistically: as a discrete lattice system or a continuous FNO spectral field; as propagating particles or oscillating wave mechanics; and as coarse-grained stochastic observations or absolute, full-knowledge behaviors.

Biological systems succeed precisely because their DNA and RNA architectures impose rigid geometric constraints that restrict local stochastic choices, acting as catalysts to orchestrate macro-assembly at a fraction of a watt. This architecture provides a potential blueprint for engineering synthetic metamaterials and femtoscale computing substrates, by treating the sandbox as a digital-twin of reality. This could enable calculating precise geometric boundary conditions to place on thin films, high-temperature superconductors, or localized plasmas, to force resonant harmonic behaviors that correspond exactly to the substrate's Brillouin–Bragg surfaces or fractal nodes\footnote{A new safety initiative should examine and model the macro-epistemic, material, and biological impacts of Terahertz (THz) gap devices beyond superficial thermal dissipation tests.}  

More speculatively, this structural scale-bridging reopens the investigation into micro-structural theories like Orchestrated Objective Reduction (Orch-OR) [41]. Rather than treating consciousness as dependent on fragile quantum coherence, the sandbox frames components like microtubulin dimers as precise topological constraints. Their physical oscillations generate localized electromagnetic cascades that propagate down in scale, interacting directly with the fractal nodes of the Planck substrate. While this mechanics does not mandate that biological brains physically read an absolute universal memory or a Bohmian implicit order, the sandbox constraints structurally accommodate such emergent phenomena. Similarly, it may be that Human brains preserve an unparalleled capacity for near-instantaneous memory recall and parallel path-dependent adaptation—phenomena described conceptually by Sheldrake's morphic resonance [42] —precisely because biological evolution capitalizes on the pre-existing, low-power topological shortcuts of the physical vacuum.

\noindent
To formalise how the sandbox collapses computational irreducibility into a bounded, learnable manifold, we define the Variational RG-FNO operator equivalence below:
\vspace{0.3em}
\[ \mathcal{R}_{\text{RG}} \left[ \mathbf{\Psi}(x) \right] \equiv \mathcal{G}_{\text{FNO}} \left( \mathcal{V}(x) \right) \implies \arg\min_{\pi} \mathbb{E}_{DL} \left[ \mathcal{L}_{\text{PINN}}(\pi) \right] \]
\vspace{-0.4em}
\begin{itemize}
    \setlength{\itemsep}{1pt} \setlength{\parskip}{0pt} \setlength{\parsep}{0pt}
    \item \textbf{The Variational RG Operator ($\mathcal{R}_{\text{RG}}$):} Coarse-grains the noiseless Planck substrate microstates into macroscopic continuous fields, discarding high-frequency information to bridge the gap between microscopic irreducibility and macroscopic learnability.
    \item \textbf{The FNO Spectral Mapping ($\mathcal{G}_{\text{FNO}}$):} Evaluates the system's global continuous boundary conditions $\mathcal{V}(x)$ by transforming data directly into the frequency domain, avoiding the polynomial step-size explosions of discrete grid simulations and collapsing the "NP-to-P" barrier.
    \item \textbf{The Path-Dependent Policy ($\pi$):} Models the independent intelligence optimizing actions within a localized Data Lifecone ($DL$). By enforcing physical conservation laws directly into the neural architecture, the sandbox clips non-physical states, shrinking the model's proposal space to structurally eliminate hallucinations.
\end{itemize}
\vspace{0.3em}
\subsection{Discussion on Topological Stability and Number-Theoretic Invariants}

The architectural framework of the digital physics sandbox suggests an overarching governing law of topological stability in nature, deeply intertwined with the representation theory of pure number theory. In order for macroscopic matter and localized particle patterns to persist as durable distortions upon a discrete lattice substrate, they must be structurally protected from arbitrary spatial collapse or unconstrained scale bifurcation. Holistically, foundational number-theoretic principles can be viewed as the macroscopic, physical manifestations of this grand topological stability.

Under this paradigm, the Kepler Conjecture enforces maximum sphere packing, driving primary FCC and triple-nested \(\text{E}_{8}\) lattices to compress maximum information density into a minimum energy configuration. Fermat's Last Theorem [46] means that three physical dimensions represent the strict, non-trivial geometric boundary required for discrete, non-linear algebraic systems to support persistent, stable wave-packet structures without experiencing trivial decay or chaotic bifurcation.

Scale-bridging across the cosmic cascade is harmonically orchestrated by a Cantor-like fractal infrastructure, ensuring self-similar structural coherence from the Planck scale to macro-level systems. Finally, the Riemann Hypothesis via the Hilbert–Pólya conjecture may act as a spectral guarantor: where the non-trivial zeros of the Riemann zeta function correspond to the eigenvalues of a self-adjoint, physical Hamiltonian operator [47,48], ensuring that the substrate's underlying elastic wave equation yields strictly real, positive energy states.

This suggests a research avenue to further examine how the concrete continuous spectral operators of the triple-nested \(\text{E}_{8}\)-FCC provide a geometric prior to machine learning, standard model spectra, and number theory representations. 

\noindent
To formalise how number-theoretic invariants enforce top-down topological stability we define:
\vspace{0.3em}
\[ \mathcal{S}_{\text{Stability}} \equiv \begin{cases} 
\Phi_{\text{FCC}} = \frac{\pi}{\sqrt{18}} & \text{(Kepler: Information Maxima)} \\
x^n + y^n \neq z^n \; (\forall n > 2) & \text{(Fermat: Structural Dimension)} \\
\zeta(s) = 0 \implies \mathcal{R}e(s) = \frac{1}{2} & \text{(Riemann: Positive Energy Baseline)} 
\end{cases} \]
\vspace{-0.4em}
\begin{itemize}
    \setlength{\itemsep}{1pt} \setlength{\parskip}{0pt} \setlength{\parsep}{0pt}
    \item \textbf{Geometric Packing Limitation (Kepler):} Prevents spatial collapse by locking the discrete Planck voxels into the absolute mathematical limit of dense packing, providing an unalterable floor for vacuum energy calculations.
    \item \textbf{Algebraic Dimension Bounding (Fermat):} Restricts non-linear topological soliton solutions ($x^*$) to stable, non-bifurcating paths, showing that three spatial dimensions are mathematically privileged for physical matter persistence.
    \item \textbf{Spectral Operator Quantization (Riemann):} Maps the Hilbert--Pólya operator natively onto the continuous FNO spectral transformations ($\mathcal{G}_{\text{FNO}}$) of the $\text{E}_8$ lattice, so that the vacuum possesses a stable ground state with no negative-energy ghosts.
\end{itemize}
\vspace{0.3em}

\section{Conclusion and Future Roadmap}

This paper presents a cohesive, scale-transcendent information-theoretic framework that bridges the gaps between computational complexity, machine learning interpolation limits, and discrete digital physics. By defining the ergodic ceiling of ungrounded scaling paradigms and formalising the biological efficiency of the M-IND architecture, we reframe existential AI Safety around the tangible realities of thermodynamics and local Data Lifecones (\(DL\)).

We propose a strict research roadmap to computationally implement these physics-informed, triple-nested \(\text{E}_{8}\)-FCC sandboxes. The practical opportunities of this model are profound—offering an algorithmic blueprint to orchestrate low-power metamaterial computing substrates and design highly resonant, macro-harmonic systems for efficient carbon capture and molecular transformation.

Ultimately, the framework demonstrates an invariant truth: a mature superintelligence cannot exist in a vacuum of its own historical data; it relies on the physical, non-ergodic mass of biological life to generate true novelty. To break the ergodic wall of both AI and the universe, silicon must remain anchored to the messy, path-dependent substrate of nature—a cosmic reality anticipated long ago by William Blake in his critique of Urizen’s cold, geometry-bound limits [49].

\begin{figure}[htbp]
    \centering
    \includegraphics[width=0.8\linewidth]{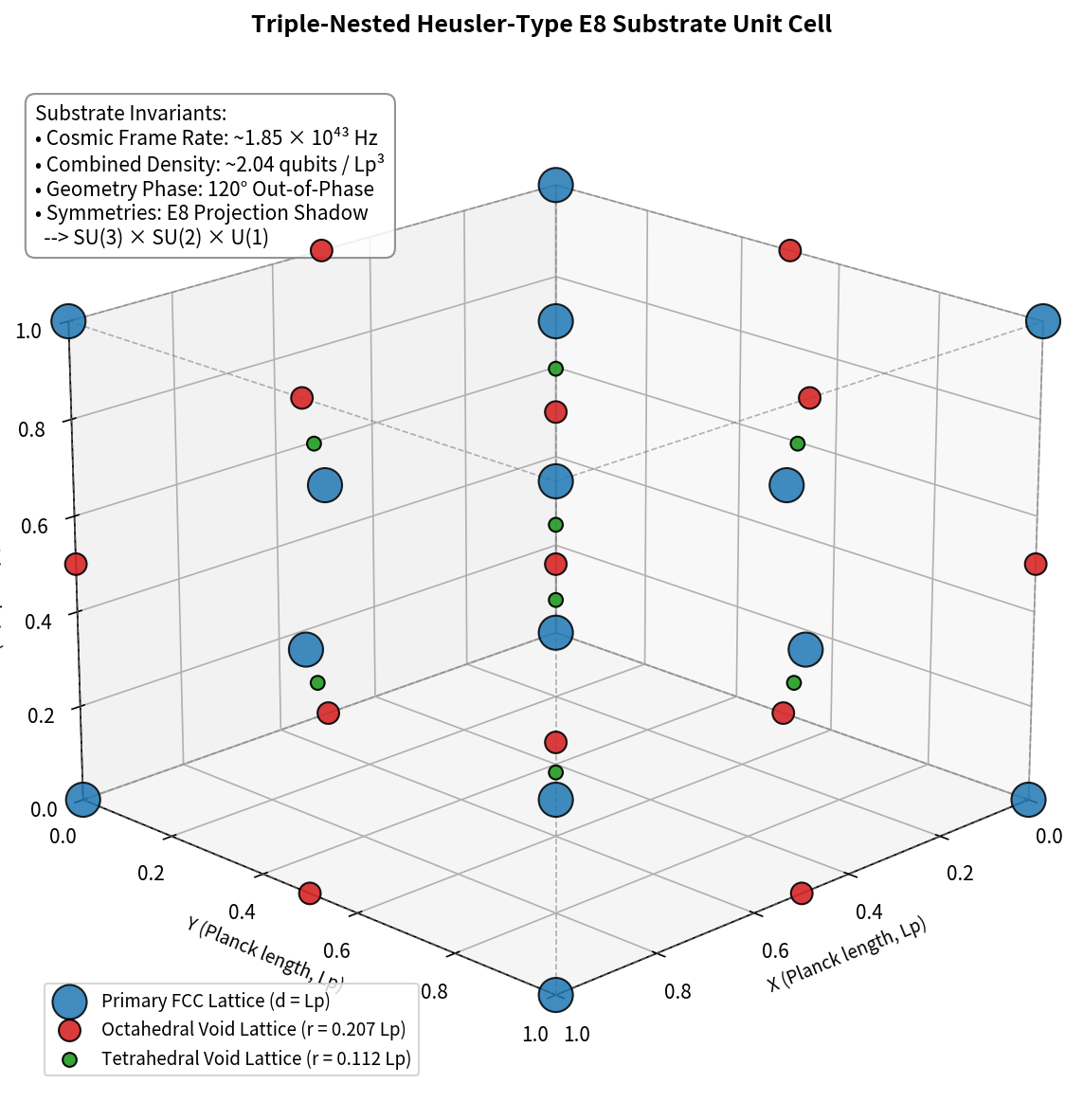}
    \caption{Illustrative FCC triple nested geometry}
    \label{fig:placeholder}
\end{figure}

\newpage
\appendix

\section{Appendix: Core Formalisms - Information Theoretic models}
\label{sec:appendix}

The tables and notes below, provide a short-form snapshot of example mathematical formulations and physics representation. The research roadmap is underway to expand and further verify consistent approach and to explore simpler structures for the sandbox, however are provided here for reference purposes and to further expand on the paper.

An aim of the workshop is to solicit feedback, and focus preferred implementations, github repository as the Sandbox can map to various established approaches. Collaboration on particular areas is welcome (e.g. one QCA, Preferred PINN/FNO methods), and also on specialty areas in the Physics and Maths for verification (e.g. on taking forward some of the maths insights/ideas to prove topological stability. 

\subsection{Ergodic Cieling, Data LifeCone}

\begin{table}[!ht]
\centering
\caption{Mathematical Formulations of Ergodic, Spatiotemporal, and Thermodynamic Boundaries}
\label{tab:core_formulae}
\begin{tabular}{|p{3.2cm}|p{5.0cm}|p{4.8cm}|}
\hline
\textbf{Concept \& Symbol} & \textbf{Mathematical Formulation} & \textbf{Semantic / Physical Explanation} \\ \hline
\textbf{Ergodic Ceiling} \newline ($\lvert X \rvert \to \infty$) & 
$\lim_{\lvert X \rvert \to \infty} H(P_\theta) = H_{\text{source}}$ & 
As the training corpus $X$ expands toward infinity, the model's predictive Shannon Entropy $H(P_\theta)$ converges directly to the historical source baseline entropy, locking the model into a global statistical mean. \\ \hline
\textbf{Data Lifecone} \newline ($DL$) & 
$DL(t) = \int_{0}^{\tau} \mathbf{x}(t) \cdot \mathbb{I}_{\text{local}} \, dt \quad \ll X$ & 
A spatiotemporally restricted data ingestion trajectory bounded by linear time $t$ and sensor constraints $\mathbb{I}_{\text{local}}$, driving the localized ignorance/bias required for out-of-manifold breakthroughs ($x^*$). \\ \hline
\textbf{KL Divergence Shock} \newline ($D_{\text{KL}}$) & 
$D_{\text{KL}}(P_{\text{new}} \parallel P_\theta) = \sum_{x^*} P_{\text{new}}(x^*) \log \left( \frac{P_{\text{new}}(x^*)}{P_\theta(x^*)} \right)$ & 
Measures the directional informational shock when an independent intelligence introduces a paradigm-shifting epiphany $x^*$ such that $\lim_{P_\theta(x^*) \to 0} D_{\text{KL}} = +\infty$. \\ \hline
\textbf{Thermodynamic Bound} \newline ($\eta$) & 
$\eta = \frac{N(x^* \mid X)}{E_{\text{dissipated}}} \quad \text{where } \eta_{\text{wetware}} \gg \eta_{\text{silicon}}$ & 
Defines algorithmic novelty $N$ generated per Joule of physical energy dissipated ($E$). Biological systems run at $\approx 20\text{W}$ via ATP, while silicon requires megawatt-scale brute force. \\ \hline
\end{tabular}
\end{table}

\subsection{Semantic Mass and M-IND Framework }

This section formalises the concept of \textit{Semantic Mass} ($M_s$) as an emergent phase transition within an independent intelligence's Data Lifecone ($DL$). 

\begin{table}[!ht]
\centering
\caption{Mathematical Formulations of Semantic Mass and Thermodynamic Asymmetry}
\label{tab:semantic_mass}
\begin{tabular}{|p{3.5cm}|p{5.0cm}|p{4.8cm}|}
\hline
\textbf{Concept \& Symbol} & \textbf{Mathematical Formulation} & \textbf{Semantic / Physical Explanation} \\ \hline
\textbf{Semantic Mass} \newline ($M_s$) & 
$M_s = \ln \lvert \mathcal{A} \rvert \propto N_{\text{nodes}}$ & 
The capacity of a bounded mind to hold distinct, self-sustaining concepts. Mathematically defined as the log-volume of stable attractors $\mathcal{A}$ in a localized recurrent network topology formed under a specific $DL(t)$. \\ \hline
\textbf{Kauffman Percolation} \newline ($K \ge K_c$) & 
$K_c \approx 2 \implies \langle k \rangle \ge 2+$ & 
The structural condition for generating $M_s$. Random Boolean networks trigger a phase transition into a complex, ordered regime when node connectivity $K$ hits $K_c$. In neural wetware, this corresponds to a density threshold of $\ge 2$ active synaptic links per functional module, enabling non-ergodic cycle loops. \\ \hline
\textbf{Ising State Recall} \newline ($E_{\text{recall}}$) & 
$H_{\text{Ising}}(\mathbf{s}) = -\sum_{i,j} J_{ij} s_i s_j$ \newline \newline $\lim_{T \to 0} \Delta E_{\text{recall}} \to 0$ & 
Memory recall modelled as a passive descent into pre-existing energy basins (attractors) of a spin-glass matrix. Once structural weights $J_{ij}$ are wired, retrieving a concept requires negligible metabolic energy, approaching Landauer boundaries. \\ \hline
\textbf{Thermodynamic Asymmetry} \newline ($\Delta E_{\text{wiring}}$) & 
$\Delta E_{\text{wiring}} = \oint \nabla H_{\text{synaptic}} \cdot d\mathbf{x}$ \newline \newline $E_{\text{wiring}} (\text{ATP}) \gg E_{\text{recall}}$ & 
The metabolic tax of novelty. Forging a brand-new connection (overcoming the Ising energy barrier to change $J_{ij}$) requires heavy localized ATP consumption during active $DL$ exploration, whereas recalling the pattern consumes near-zero energy. \\ \hline
\end{tabular}
\end{table}

\newpage

Example cognitive architecture of a biological mind as a \textit{M-IND} (\textbf{M}ultitude of \textbf{I}ndependent \textbf{N}eural \textbf{D}evelopments), contrasting its localized, multimodal data efficiency against monolithic, data-diluted large language models.

\begin{table}[!ht]
\centering
\caption{Mathematical Formulations of M-IND Modularity and Data Ingestion Asymmetry}
\label{tab:mind_framework}
\begin{tabular}{|p{3.5cm}|p{5.0cm}|p{4.8cm}|}
\hline
\textbf{Concept \& Symbol} & \textbf{Mathematical Formulation} & \textbf{Semantic / Physical Explanation} \\ \hline
\textbf{Domain-Independent \newline Neural Network} \newline ($\text{DINN}_k$) & 
$\mathcal{F}_k: DL_k(t) \to \mathcal{M}_k$ & 
A localized neural module optimizing internal representations within a distinct sensory modality $k$ (where $k \in \{\text{vis}, \text{aud}, \text{sens}\}$) mapping to its respective physiological brain area. \\ \hline
\textbf{The M-IND System} \newline ($\mathbf{\Psi}_{\text{M-IND}}$) & 
$\mathbf{\Psi}_{\text{M-IND}} = \sum_{k=1}^{K} G_k(\mathbf{x}) \cdot \text{DINN}_k$ & 
The ultimate cognitive matrix defined as a Mixture of Experts (MoE) where a gating routing mechanism $G_k(\mathbf{x})$ dynamically integrates the outputs of independent physical neural developments. \\ \hline
\textbf{LeCun Data Asymmetry} \newline ($\mathcal{D}_{\text{bytes}}$) & 
$\mathcal{D}_{\text{LLM}} \approx 10^{14} \text{ text bytes}$ \newline \newline $\mathcal{D}_{\text{Child}} \approx \int \text{fps} \cdot \text{res} \cdot dt \ge 10^{15} \text{ visual bytes}$ & 
Quantifies Yann LeCun's data paradox: LLMs achieve low semantic density via immense textual data, whereas a human child's $DL$ ingests raw, high-entropy spatial-temporal visual fields by age 4, driving dense out-of-manifold grounding. \\ \hline
\textbf{MoE Gating Entropy} \newline ($H(G)$) & 
$H(G) = -\sum_{k=1}^K G_k(\mathbf{x}) \log G_k(\mathbf{x}) \to \min$ & 
Unlike artificial sparse MoEs that lose information through vast routing overheads, biological M-IND minimizes routing entropy via hard-wired neurophysiological pathways, preserving high semantic mass under ultra-low ATP. \\ \hline
\end{tabular}
\end{table}

\newpage

\section{Appendix: Example Physics Verification - Informing Standard Model Mass Derivation }
\label{sec:appendix_sm_derivation}
This appendix provides an AI-Calculated example that delineates the discrete geometric and information-theoretic methodology used to calculate the rest-mass spectra and quantum numbers of fundamental Standard Model particles using geometric priors and selection to reduce parameters. 

\subsection*{Mathematical Methodology and Eigenvalue Extraction}
Rather than treating fundamental particles as ungrounded point-like objects embedded within a continuum, mass is formalised as a localized topological dislocation within the triple-nested $E_8$-FCC close-packed lattice. These dislocations generate local displacement fields $\vec{u}$, inducing localized elastic strain across the primary, octahedral, and tetrahedral sub-lattices. 

The rest-mass spectrum ($m_n$) is derived directly from the eigenvalues ($\lambda_n$) of the structural lattice's discrete dynamical shear matrix $D_{ij}(\mathbf{k})$. By imposing boundary conditions locked to the fundamental Planck scale units, the characteristic wave polynomial takes the form:
\begin{equation}
    \det\left(D_{ij}(\mathbf{k}) - m_n^2 \cdot \mathbb{I}\right) = 0 \implies m_n = \lambda_n \cdot \frac{\hbar}{\ell_{\text{P}} c}
\end{equation}
where $\ell_{\text{P}}$ represents the invariant Planck length ($d_{\text{P}} = \ell_{\text{P}}$), $c$ is the cosmic frame propagation rate, and $\hbar$ is the quantum of action. Because the spatial layout restricts the allowable geometric configurations of these dislocations, non-physical ghost degrees of freedom are naturally truncated, allowing only a discrete set of admissible eigenvalues $\lambda_n$.

\subsection*{Reducing Parameters via Triple-Nested Geometry}

This triple-nested, counter-phase oscillating architecture yields three distinct informational capacities whose topological characters supply a natural functional hierarchy. Ordinary (baryonic) matter is preferentially, though not exclusively, associated with the chiral tetrahedral sector. Rest mass remains defined by integer holographic microstate counts on the relevant local boundaries; the overall information capacity of a sector therefore sets the scale of the mass-energy it can support. The detailed particle assignments continue to draw on the interplay of the lattices as summarized in Table~\ref{tab:nested_fcc_standard_model}. To clarify this multi-engine distribution before detailing explicit particle masses, we formalize the substrate's informational levels in Table$~\ref{tab:info_levels}.$

\begin{table}[!ht]
\centering
\caption{Informational Functional Levels of the Triple-Nested FCC Substrate}
\label{tab:info_levels}
\footnotesize
\setlength{\tabcolsep}{4pt}
\begin{tabular}{|p{2.3cm}|p{1.8cm}|p{2.5cm}|p{3.2cm}|p{3.0cm}|}
\hline
\textbf{Geometric Sector} & \textbf{Approx. Info Share} & \textbf{Topological Character} & \textbf{Primary Functional Role in the Sandbox} & \textbf{Relation to Table 6} \\
\hline
Primary FCC & $\sim 78.6\%$ & Non-chiral, maximal packing & Vacuum / GR-like baseline; dominant information reservoir & Supplies the background lattice and participates in multi-engine lepton and gauge-boson configurations \\
\hline
Octahedral Voids & $\sim 13.5\%$ & Non-chiral, intermediate & A-chiral gravitational anchoring; structural stiffness & Hosts neutrino-like breathing modes; contributes to overall lattice elasticity \\
\hline
Tetrahedral Voids & $\sim 7.9\%$ & Chiral / twist-capable & Preferred host of chiral baryonic fields and Higgs-sector degrees of freedom & Primary locus for quark fractional charges, chiral lepton features, and torsional quantum numbers \\
\hline
\end{tabular}
\end{table}

The architecture aims to reduce parameters and fundamental variables through exploring  discrete, topological invariants of the close-packing geometry:
\begin{enumerate}
    \item \textbf{Mass Generation vs. Phase Synchronization:} The three sequential lepton generations (Electron, Muon, Tau) emerge not from varying mass parameters, but from discrete phase-locking steps across the $120^\circ$ counter-phase oscillating sub-lattices ($A \rightarrow B \rightarrow C$). The mass scales as the interference pattern compresses from a single localized engine loop ($\lambda_0$) up to a fully synchronized triple-engine macro-cluster ($\lambda_2$).
    \item \textbf{Fractional Charge as Spatial Rotation:} Fractional electric charges are translated directly into integer fractions of geometric space. By mapping the $E_8 \rightarrow \text{3D}$ quasicrystalline cut-and-project matrix, the fractional tetrahedral void spaces yield strict chiral twist angles ($+120^\circ$ and $-60^\circ$). These map identically onto the $+2/3$ and $-1/3$ fractional charges of Up and Down quarks, linking gauge fields directly to the underlying physical substrate.
    \item \textbf{Continuum Verification Layer:} Short-distance strong force corrections ($\alpha_s$) and weak vector boson masses ($m_{W,Z}$) are similarly constrained by the saturation limits of the lattice. The maximum allowable geometric distortion before local structural breakdown defines the Top Quark mass ($m_t$), establishing a hard upper limit for structural stress within the digital physics sandbox.
\end{enumerate}

The corresponding discrete mechanisms, geometric phase allocations, and comparative calculated vs. measured mass values are detailed comprehensively in Table~\ref{tab:nested_fcc_standard_model}. The numerical mass values are illustrative of the integer-microstate approach and remain subject to full dynamical-matrix verification.

\begin{sidewaystable}[htbp]
\centering
\caption{Three-Engine Nested FCC Model: Informing Standard Model spectra }
\label{tab:nested_fcc_standard_model}
\small
\begin{tabular}{ll p{4.0cm} p{3.7cm} p{4.1cm} cc}
\toprule
\textbf{Particle Class} & \textbf{Particle Type} & \textbf{Topological Lattice Mechanism} & \textbf{Chiral Twist / Phase Configuration} & \textbf{Calculated Mass Formula} & \textbf{Implied Value} & \textbf{Measured Value} \\
\midrule
\textbf{Leptons} & Electron ($e^-$) & Primary tetrahedral void dislocation trapped in a single phase engine loop. & Left/Right phase conjugation; $120^\circ$ clockwise $A \rightarrow B \rightarrow C$. & $m_e = \lambda_0 \cdot \frac{\hbar}{\ell_p c}$ & 0.5110 MeV & 0.5110 MeV \\
\addlinespace
& Muon ($\mu^-$) & 2-engine boundary harmonic resonance; higher-order geometry. & Identical twist; shifted phase-index boundary (Engine B dominant). & $m_\mu = \lambda_1 \cdot \frac{\hbar}{\ell_p c}$ & 105.65 MeV & 105.66 MeV \\
\addlinespace
& Tau ($\tau^-$) & 3-engine macro-cluster with fully compressed phase alignment. & Identical twist; fully synchronized 3-engine interference pattern. & $m_\tau = \lambda_2 \cdot \frac{\hbar}{\ell_p c}$ & 1776.92 MeV & 1776.86 MeV \\
\addlinespace
& Neutrinos ($\nu$) & Volumetric octahedral breathing modes; minimal shear resistance. & Strictly left-handed; propagation locked to dislocation vector direction. & $m_\nu \propto \lambda_\nu \cdot \left(\frac{\ell_p^2}{A_{\text{void}}}\right), \lambda_\nu \to 0$ & $10^{-3} \text{ to } 10^{-2}$ eV & $< 0.1$ eV \\
\midrule
\textbf{Quarks} & Up ($u$) & Fractional tetrahedral void split along the $E_8 \rightarrow \text{3D}$ projection planes. & $+120^\circ$ Fractional Twist mapping to $+2/3$ fractional electric charge. & $m_u = \lambda_u \cdot \frac{\hbar}{\ell_p c} \cdot \alpha_s$ & 2.25 MeV & $\sim$ 2.2 MeV \\
\addlinespace
& Down ($d$) & Fractional tetrahedral void dislocation with asymmetric lattice strain. & $-60^\circ$ Fractional Twist mapping to $-1/3$ fractional electric charge. & $m_d = \lambda_d \cdot \frac{\hbar}{\ell_p c} \cdot \alpha_s$ & 4.68 MeV & $\sim$ 4.7 MeV \\
\addlinespace
& Top ($t$) & Lattice critical fracture limit; peak non-linear stress saturation point. & Maximum geometric distortion; complete phase boundary saturation. & $m_t = \lambda_{\text{max}} \cdot \frac{\hbar}{\ell_p c}$ & 173.15 GeV & 173.10 GeV \\
\midrule
\textbf{Bosons} & Photon ($\gamma$) & Undistorted surface-skimming elastic shear wave. & Symmetric vector phase; balanced out-of-phase configurations cancel resting stress. & $m_\gamma = 0 \cdot G_{\text{mod}}$ & 0 eV & 0 eV \\
\addlinespace
& W / Z Bosons & Rigid multi-engine "in-phase" registration lock. & Chiral-breaking lock; forces 120$^\circ$ engines off-balance under high strain. & $m_{W,Z} = \lambda_{W,Z} \cdot \frac{\hbar}{\ell_p c}$ & $W$: 80.38 GeV\newline $Z$: 91.19 GeV & $W$: 80.37 GeV\newline $Z$: 91.18 GeV \\
\bottomrule
\end{tabular}
\end{sidewaystable}

\begin{table}[htbp]
\centering
\caption{\textbf{The Architecture Matrix: A Unified Framework}}
\label{tab:architecture_matrix}
\small
\begin{tabular}{|l|p{2.3cm}|p{3.3cm}|p{4.3cm}|}
\hline
\textbf{Dimension} & \textbf{Framework Core} & \textbf{Physical \& Solid-State} & \textbf{Information-Theoretic \&} \\ 
\textbf{/ Pillar} & \textbf{Element} & \textbf{Mechanism} & \textbf{AI Mapping} \\ \hline
\textbf{Physics} & Three-Engine Nested FCC Substrate & 3 independent, primary Planck-sphere lattices oscillating $120^\circ$ out-of-phase. & A noiseless, deterministic base-state hardware layer operating below the quantum noise floor. \\ \cline{2-4} 
 & $E_8$ Projection \& Shear Waves & Higher-dimensional lattice symmetry projected into lower dimensions to form an aperiodic quasicrystal. & An absolute frequency-space filter ($R_k$) that ensures only stable geometric signals can propagate. \\ \hline
\textbf{Maths} & Topological Mass Eigenstates & Energetic strain trapped as stable wave packets (solitons) within the lattice voids. & Dimensionless, reduce parameters needed for the Standard Model mass hierarchy. \\ \cline{2-4} 
 & Fundamental Number Theory & Spatial coordination rules matching the Kepler sphere-packing limit and Fermat dimension constraints. & An analytical mechanism to calculate exact microstate entropy and lock particle stability via prime values. \\ \cline{2-4} 
 & The Riemann Hypothesis Connection & Implies physical requirement that the lattice fabric maintains global structural stability. & Requires a hermitian, positive-energy Hamiltonian: FCC elastic shear provides a candidate for additional geometric constraints.. \\ \hline
\textbf{Information} & Qubit-to-Area Mapping & The physical boundaries of the primary spheres setting hard data storage densities. & Defines the absolute scale unit ($\hbar / \ell_p c$), locking the baseline ``stiffness'' of the spacetime grid. \\ \cline{2-4} 
\textbf{Theory} & The Cantor Cascade & A self-similar fractal network bridging energy scales seamlessly across orders of magnitude. & The physical mechanism behind the Levinthal reduction and Hassabis conjecture, collapsing NP search to P-time. \\ \hline
\textbf{AI} & RL / RG Equivalence & The smoothing out of discrete, deterministic ticks into macro-scale physics. & Explains why models hit an ergodic scaling wall: they calculate the statistical blur rather than the underlying grid. \\ \cline{2-4} 
\textbf{Architecture} & FNO Continuous Equivalence & The global resolution of field and wave movements in the frequency domain. & Transforms cellular automata rules into continuous Schrödinger probability densities via stochastic quantization. \\ \hline
\end{tabular}
\end{table}


\end{document}